\documentclass{article}

\usepackage{natbib}
\usepackage{fullpage}

\usepackage[utf8]{inputenc} 
\usepackage[T1]{fontenc}    
\usepackage{xcolor}
\usepackage{hyperref}       
\definecolor{myblue}{rgb}{0,0.2,0.8}
\hypersetup{ %
pdftitle={},
pdfauthor={},
pdfsubject={},
pdfkeywords={},
pdfborder=0 0 0,
pdfpagemode=UseNone,
colorlinks=true,
linkcolor=myblue,
citecolor=myblue,
filecolor=myblue,
urlcolor=myblue,
pdfview=FitH}
\usepackage{authblk}
\usepackage{url}            
\usepackage{booktabs}       
\usepackage{amsfonts}       
\usepackage{nicefrac}       
\usepackage{microtype}      

\usepackage{graphicx,color}
\usepackage{xspace}
\usepackage[varqu,varl,var0,scaled=0.97]{inconsolata}
\usepackage{caption}
\usepackage{overpic}
\usepackage{wrapfig}
\usepackage{tcolorbox}
\usepackage{enumitem}
\usepackage{threeparttable}

\usepackage{algorithm}
\usepackage{algpseudocode}
\usepackage{listings}
\usepackage{color}
\usepackage{lipsum}

\usepackage{subfigure}

\definecolor{dkgreen}{rgb}{0,0.6,0}
\definecolor{gray}{rgb}{0.5,0.5,0.5}
\definecolor{mauve}{rgb}{0.58,0,0.82}

\newcommand{\name}{REPAIR\xspace}

\newcommand{\etal}{\emph{et~al.}\xspace}
\newcommand{\eg}{\emph{e.g.},\xspace}
\newcommand{\ie}{\emph{i.e.},\xspace}

\usepackage{amsmath,amsfonts,bm}

\def\Figref#1{Figure~\ref{#1}}

\def\Secref#1{Section~\ref{#1}}

\def\eqref#1{equation~\ref{#1}}
\def\Eqref#1{Equation~\ref{#1}}

\def\1{\bm{1}}

\DeclareMathAlphabet{\mathsfit}{\encodingdefault}{\sfdefault}{m}{sl}
\SetMathAlphabet{\mathsfit}{bold}{\encodingdefault}{\sfdefault}{bx}{n}

\newcommand{\E}{\mathbb{E}}

\newcommand{\R}{\mathbb{R}}

\newcommand{\Var}{\mathrm{Var}}

\newcommand{\Cov}{\mathrm{Cov}}

\newtheorem{prop}{Conjecture}
\newcommand{\remcomment}[1]{}
\RequirePackage{xr-hyper}

\title{\bf{REPAIR: REnormalizing Permuted Activations for Interpolation Repair}}
\author[1]{Keller Jordan\footnote{This collaboration was facilitated by \href{https://mlcollective.org/}{ML Collective}}}
\author[2]{Hanie Sedghi}
\author[3]{Olga Saukh}
\author[3]{Rahim Entezari}
\author[2]{Behnam Neyshabur}
\affil[1]{Hive AI}
\affil[2]{Google Research}
\affil[3]{TU Graz / CSH Vienna}
\begin{document}
\date{}
\maketitle

\begin{abstract}
In this paper we look into the conjecture of Entezari \etal~(2021) which states that if the permutation invariance of neural networks is taken into account, then there is likely no loss barrier to the linear interpolation between SGD solutions. First, we observe that neuron alignment methods alone are insufficient to establish low-barrier linear connectivity between SGD solutions due to a phenomenon we call \emph{variance collapse}: interpolated deep networks suffer a collapse in the variance of their activations, causing poor performance. Next, we propose \name (REnormalizing Permuted Activations for Interpolation Repair) which mitigates variance collapse by rescaling the preactivations of such interpolated networks. We explore the interaction between our method and the choice of normalization layer, network width, and depth, and demonstrate that using \name on top of neuron alignment methods leads to 60\%-100\% relative barrier reduction across a wide variety of architecture families and tasks. In particular, we report a 74\% barrier reduction for ResNet50 on ImageNet and 90\% barrier reduction for ResNet18 on CIFAR10. Our code is available at \url{https://github.com/KellerJordan/REPAIR}.
\end{abstract}

\section{Introduction}
Training a neural network corresponds to optimizing a highly non-linear function by navigating a complex loss landscape with numerous minima, symmetries and saddles~\citep{zhang2016understanding,  keskar2017largebatch, draxler2018essentially,csimcsek2021geometry}. Overparameterization is one of the reasons behind the abundance of minima leading to different functions that behave similarly on the training data~\citep{Neyshabur2017exploring, nguyen2018loss, li2018over, liu2020loss}. Another reason is the existence of permutation and scaling invariances which lead to functionally identical minima that differ in the weight space~\citep{brea2019weight, entezari2021role}.
Due to the relationship of the loss landscape with generalization and optimization, a large body of recent works~\citep{li2017visualizing, mei2018mean, geiger2019jamming, nguyen2018loss, fort2019deep, csimcsek2021geometry,juneja2022linear} study the loss landscape of deep neural networks with the goal of navigating the optimizer to a region with desired properties, \eg with respect to flatness around the SGD solution~\citep{baldassi2020shaping, pittorino2020entropic}.

Early work conjectured the existence of a \textit{non-linear} path of non-increasing loss between solutions found by SGD~\citep{freeman2016topology,draxler2018essentially} and empirically showed how to find it~\citep{garipov2018loss,tatro2020optimizing,pittorino2022deep}.
\textit{Linear} paths between SGD solutions are investigated by \citet{nagarajan2019uniform}, who studied MLPs trained from the same initialization on disjoint subsets of MNIST.
\citet{frankle2020linear} studied linear mode connectivity at a larger scale, and focused on different
samples of SGD noise rather than disjoint samples of data. \citet{frankle2020linear} also showed the correspondence between linear mode connectivity and Lottery Ticket Hypothesis ~\citep{frankle2019lottery}~\ie solutions that are linearly connected with no barrier have the same lottery ticket.
Recently, \citet{entezari2021role} conjectured the existence of such a \textit{linear} path between SGD solutions if the permutation invariance of neural networks' weight space is taken into account. That is, with high probability over SGD solutions, for each pair of trained networks A and B there exists a permutation of the hidden units in each layer of B such that the linear path between A and the permuted network B (B') is of non-increasing loss relative to the endpoints.
This conjecture is important from both theoretical and empirical perspectives. Theoretically, it leads to a drastic simplification of the loss landscape, reducing the complexity obstacle for analyzing deep neural networks. Empirically, linear interpolation between neural network weights has become an important tool, having recently been used to set state of the art accuracy on ImageNet \citep{wortsman2022model}, improve robustness of finetuned models \citep{wortsman2022robust,ilharco2022patching}, build effective weight-space model ensembles~\citep{,izmailov2019averaging, frankle2020linear, hao2022swarevisited}, and constructively merge models trained on separate data splits \citep{wang2020federated,ainsworth2022git}. Therefore, any improvements toward reducing the obstacles to interpolation between trained models has the potential to lead to empirical progress in the above areas.

Prior and concurrent works on linear interpolation \citep{singh2020model,entezari2021role,ainsworth2022git} have focused on improving the algorithms used to bring the hidden units of two networks into alignment, in order to reduce the barrier to interpolation between them. \citet{singh2020model} develop a strong optimal transport-based method which allows linear interpolation between a pair of ResNet18~\citep{he2016deep} networks such that the minimum accuracy attained along the path is 77\%. This constitutes a ``barrier'' of 16\% relative to the original endpoint networks which achieve over 93\% accuracy on the CIFAR-10 test set. \citet{entezari2021role} use an approach based on simulated annealing \citep{zhan2016SA} in order to find permutations such that wide multi-layer perceptrons (MLPs)~\citep{rosenblatt1958perceptron} trained on MNIST~\citep{lecun1998mnist} can be linearly interpolated with a barrier of nearly zero. \citet{ainsworth2022git} make the first demonstration of zero-barrier connectivity between wide ResNets trained on CIFAR-10 by replacing\footnote{See the code release, \url{https://github.com/samuela/git-re-basin/blob/main/src/resnet20.py\#L18}} the standard Batch Normalization \citep{ioffe2015batch} layers with Layer Normalization \citep{ba2016layer}, and develop several novel alignment methods. Further discussion of related work can be found in Appendix~\ref{app:further_related}. Given this context, in this paper we are interested in understanding why alignment of the endpoint networks alone has so far been insufficient to reach low-barrier linear connectivity between standard deep convolutional networks.

\begin{figure}
    \centering
    \subfigure{\includegraphics[height=3.6cm]{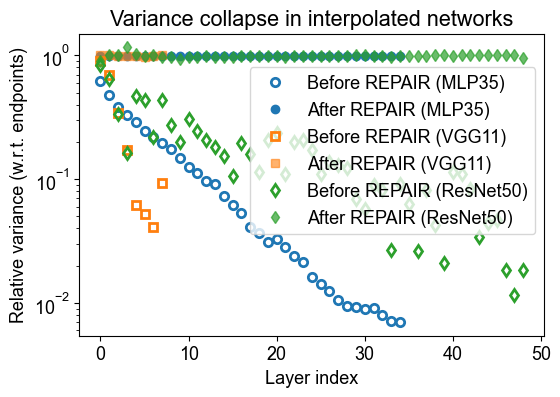}}
    \subfigure{\includegraphics[height=3.6cm]{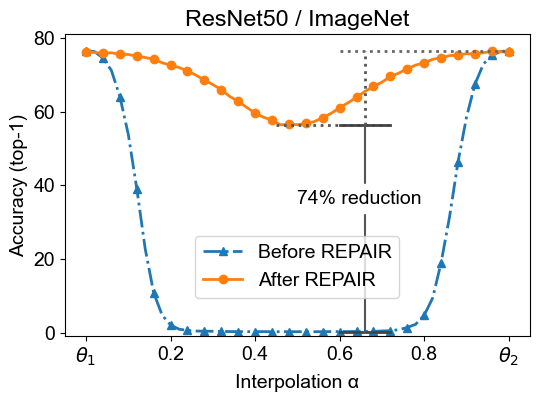}}
    \subfigure{\includegraphics[height=3.6cm]{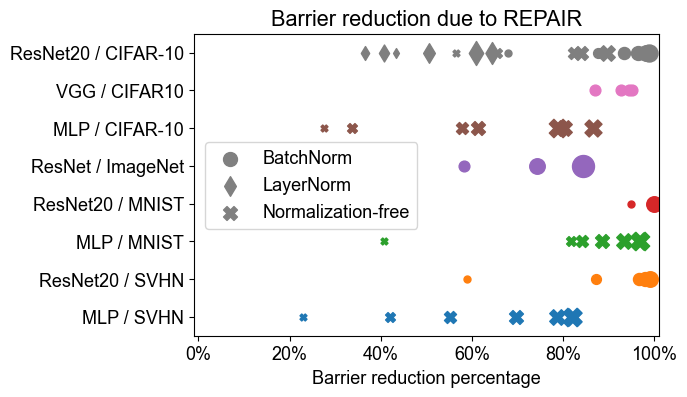}}
    \caption{\small{\textbf{\name improves the performance of interpolated networks by mitigating variance collapse.} In each experiment, we interpolate between the weights of two independently trained networks whose hidden units have been aligned using the method described in Section~\ref{sec:alignment}. We then compare the interpolated network before and after applying our correction method \name. \textbf{Left:} The variance of activations in interpolated networks progressively collapses. We report the average variance across each layer, normalized by that of the corresponding layer in the original endpoint networks. \name is designed to correct this phenomenon. \textbf{Middle:} \name reduces the barrier to linear interpolation between aligned ResNet50s independently trained on ImageNet by 74\% (from 76\% to 20\%). \textbf{Right:} \name reduces the interpolation barrier across many choices of architecture, training dataset, and normalization layer. For each architecture/dataset pair we vary the network width; larger markers indicate wider networks.
    }}
    \label{fig:fig1}
\end{figure}

\paragraph{Contributions} In this work we focus on understanding the source of the poor performance of standard deep networks (ResNet18, VGG11) whose weights have been linearly interpolated from between pairs of networks with aligned neurons. Our contributions are as follows:
\begin{itemize}
\item We find that such interpolated networks suffer from a phenomenon of \emph{variance collapse} in which their hidden units have significantly smaller activation variance compared to the corresponding units of the original networks from which they were interpolated. We further identify and explain the reason behind this variance collapse.  (Figure~\ref{fig:fig1} (left) and Section~\ref{sec:variancecollapse}).
\item We propose \name (REnormalizing Permuted Activations for Interpolation Repair), a method that corrects variance collapse by rescaling hidden units in the interpolated network such that their statistics match those of the original networks. (Section~\ref{sec:correction}). 
\item We demonstrate that applying \name to such interpolated networks leads to significant barrier reductions across a wide variety of architectures, datasets, normalization techniques, and network width/depth (Section~\ref{sec:results} and Figure~\ref{fig:fig1} (middle and right)).
\end{itemize}

\section{Preliminaries}
In this section we give preliminary definitions and algorithms which will be used throughout the paper. In summary, we first present the notion of interpolation between networks, and define the linear interpolation barrier. We then discuss the existence of permutation symmetries in weight space, and present algorithms used to locate such permutations in order to bring the hidden units of two networks into alignment.

\subsection{Linear interpolation of neural networks}
We consider the problem of interpolating between independently trained neural networks. That is, if we let $\theta_1, \theta_2$ be the weight vectors of two such networks, then we are interested in networks whose weights are of the form $\theta_\alpha = (1 - \alpha)\,\theta_1 + \alpha\,\theta_2$ for $0 < \alpha < 1$. We refer to such networks $\theta_\alpha$ as \emph{interpolated networks}, and to $\theta_1, \theta_2$ as the \emph{endpoint networks}.

The \emph{loss barrier} $B(\theta_1, \theta_2)$ between a pair of networks is defined~\citep{entezari2021role} as the maximum increase in loss along the linear path between $\theta_1$ and $\theta_2$, relative to the corresponding convex combination of the two endpoint losses. In notation:
\begin{equation}
\begin{split}
\label{eqn:barrier}
B(\theta_1, \theta_2) =
\sup_{\alpha \in [0, 1]} \Big[\mathcal{L}((1 - \alpha) \theta_1 + \alpha\theta_2)\Big] - 
\Big[(1 - \alpha)\mathcal{L}(\theta_1) + \alpha\mathcal{L}(\theta_2)\Big]. 
\end{split}
\end{equation}
The barrier between pairs of unmodified, independent SGD solutions $\theta_1, \theta_2$ is typically large. We next discuss permutation invariance in neural networks.

\subsection{Permutation invariance}
For typical neural architectures, the neurons in each layer can be permuted without functionally changing the network; this is known as the \emph{permutation invariance} property. In a simple feedforward network, this amounts to the observation that we can replace the $L$th weight matrix by $PW_L$ and the $(L+1)$th weight by $W_{L+1}P^{-1}$ without changing the function represented by the network. As a result, even if our two networks $\theta_1, \theta_2$ have learned a functionally identical set of neurons at each layer, it is possible for such neurons to be arbitrarily permuted or misaligned.

\citet{entezari2021role} conjecture that if permutation invariance is taken into account, then all SGD solutions for a sufficiently wide network trained on the same task become linearly mode connected, \ie have no barrier between them. For completeness we provide the formal statement of the conjecture below.

\begin{prop}\citep{entezari2021role}\label{thm:conj}
For a given neural architecture, let $\mathcal{P}$ be the set of all valid permutations of hidden units, and $P:\R^k \times \mathcal{P} \rightarrow \R^k$ be the function that applies a given permutation to a weight vector and returns the permuted version. Then for sufficiently wide networks the following holds:  There exists a set of solutions $\mathcal{S}\subseteq \R^k$ and a function $Q:\mathcal{S} \rightarrow \mathcal{P}$ such that for any $\theta_1,\theta_2\in \mathcal{S}$, we have small interpolation barrier $B(P(\theta_1,Q(\theta_1)),\theta_2) \approx 0$ and with high probability over an SGD solution $\theta$, we have $\theta\in \mathcal{S}$.
\end{prop}

\subsection{Neuron alignment algorithms} \label{sec:alignment}
A number of works have proposed methods of finding such alignments between the hidden units of a pair of neural networks. \citet{li2015convergent} propose to maximize the sum of correlations between the activations of paired neurons across a batch of training data. That is, if we let $X_{l,i}^{(0)}$ and $X_{l,i}^{(1)}$ be random variables corresponding to the activations of the $i$-th hidden units of the $l$-th layer (across a batch of training data), then \citet{li2015convergent} proposes to optimize the permutation $P_l$ to maximize the following objective:
\begin{equation}
\begin{split}
\label{eqn:matching}
\sum_i \mathrm{corr}(X_{l,i}^{(1)}, X_{l,P_l(i)}^{(2)}). 
\end{split}
\end{equation}
This amounts to a linear sum assignment problem corresponding to the matrix of correlations between pairs of hidden units in the two networks; which can be solved via the Hungarian algorithm~\citep{Kuhn2010TheHM}. Recent works have proposed alternative approaches: \citet{he2018multi} compute Hessian approximation to align functionally similar neurons, and \citet{singh2020model} develop an optimal transport-based method of soft-alignment. \citet{ainsworth2022git} compare three methods, including one based on that of \citet{li2015convergent} and two novel approaches.

\citet{tatro2020optimizing} also perform alignment based on minimizing \Eqref{eqn:matching}, in order to reduce the barrier to non-linear interpolation. Furthermore, they show using a proximal alternating minimization scheme that alignments found using this method are nearly optimal for their purposes. In this paper, we stick with what works and continue to use this alignment method which was originally introduced by \citet{li2015convergent}.

We note that for networks with residual connections, care must be taken to restrict the set of permutations such that the function represented by the network does not change. In particular, the same permutation of hidden units must be applied to all layers which feed into a single residual stream.

\begin{figure}
    \centering
    \subfigure{\includegraphics[width=.32\textwidth]{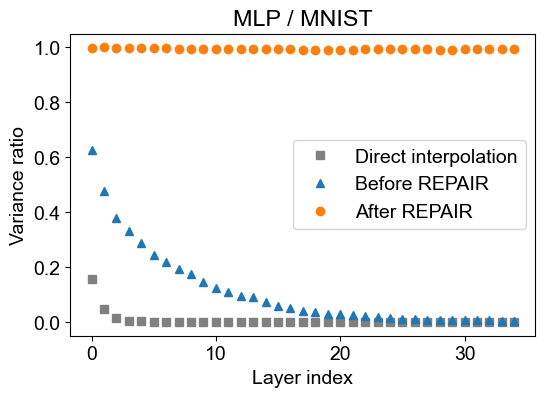}}
    \subfigure{\includegraphics[width=.32\textwidth]{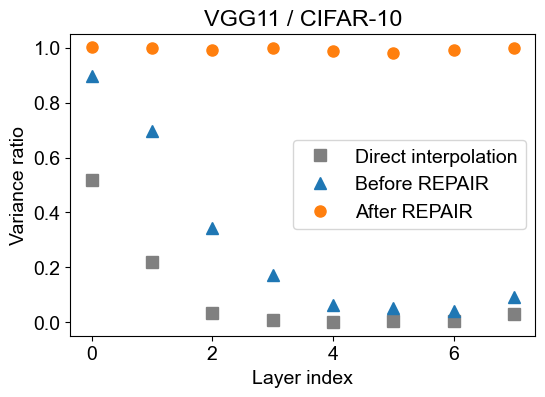}}
    \subfigure{\includegraphics[width=.32\textwidth]{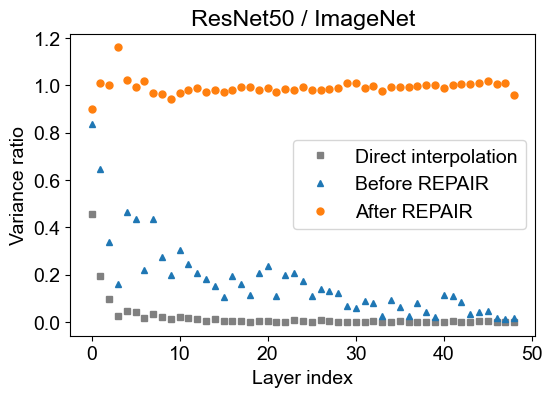}}
    \caption{\small\textbf{Variance collapse phenomenon in averaged networks.} We find that the hidden units of weight-space averaged neural networks suffer from {\it variance collapse}: as we progress through the network, variance of neuron activations reduces, with neurons in deeper layers becoming nearly constant while varying the input data. \emph{Before \name} refers to networks which are interpolated from endpoint networks whose hidden units have been aligned~\citep{li2015convergent,singh2020model,ainsworth2022git}, before our correction method \name is applied. \name is applied on top of this baseline in order to restore the internal statistics of averaged networks back to the level of the endpoint networks.}
    \label{fig:layerwise}
\end{figure}

\section{Variance collapse in interpolated networks}
\label{sec:variancecollapse}

In this section we commence our study of interpolated networks. We pick up where previous works left off:

\begin{itemize}
    \item The barrier between aligned networks decreases with width, including to nearly zero for very wide MLPs trained on MNIST. The barrier also increases sharply with depth, becoming large for MLPs or simple CNNs of more than a few layers~\citep{entezari2021role}.
    \item Even strong optimal transport-based methods, which go far beyond alignment by allowing each neuron in network A to be matched to a weighted sum of neurons in network B, are insufficient to achieve low-barrier (below 5\% test-error) connectivity between standard ResNets~\citep{singh2020model}.
\end{itemize}

We focus on understanding the source of this sharp increase in barrier between aligned networks as a function of depth observed in \citet{entezari2021role}, which we hypothesize to be the same phenomenon that causes a high barrier to interpolation for standard ResNets. 

\subsection{Identifying the problem: Variance Collapse}
What causes this rapid drop-off in the performance of interpolated networks which are deeper than a few layers? To answer this question, we investigate the internal behavior of such networks, focusing on the statistics of hidden units (\Figref{fig:layerwise}). We find that for deep MLPs, interpolated from a pair of aligned endpoint networks which both have high accuracy on the MNIST test-set, hidden units undergo a \emph{variance collapse}. That is, the variance of their activations progressively decays as we move deeper into the network, with the activations of later layers becoming nearly constant. For each layer, we quantify this decay as follows. First, we measure the variance of the activations of each neuron across a batch of training data. We then take the sum of this variance across each neuron in the layer. Finally, if we let this sum be denoted $v_\alpha, v_1, v_2$ for the interpolated and two endpoint networks, respectively, then we report the ratio $\frac{v_\alpha}{(v_1+v_2)/2}$. We compute this ratio for each layer in the network, giving a sequence of values which we report in \Figref{fig:layerwise} (left). For the set of variances of each neuron in a single layer of an interpolated ResNet18, see \Figref{fig:basicmlpbarrier} (left).

We observe that variance decays to nearly zero by the final layer of an interpolated 35-layer MLP, indicating that the activations in these last layers have become nearly constant. This effect seems to be further exacerbated when directly interpolating between unaligned networks. We repeat this experiment for VGG~\citep{simonyan2014very} and ResNet50 architectures, trained on CIFAR-10 and ImageNet respectively, and find that variance by the final layers decays by more than 10$\times$ (\Figref{fig:layerwise} (middle) and \Figref{fig:layerwise} (right)). This is a problem: if these networks have nearly constant activations in their final layers, then they will no longer even be able to differentiate between inputs.

\subsection{Why does this phenomenon occur?}
We argue that this phenomenon can be understood through the following statistical calculation. Consider a hidden unit or channel in the first layer of the interpolated network. Such a unit will be functionally equivalent to the linear interpolation between the respective units in the endpoint networks. That is, if we represent the unit's preactivation by $X_\alpha$ in the interpolated network, and $X_1, X_2$ in the two endpoint networks (as random variables over the input data distribution), then the equality $X_\alpha = (1 - \alpha) X_1 + \alpha X_2$ holds. We will argue that the variance of $X_\alpha$ is typically reduced as compared to that of $X_1$ or $X_2$.

If the two endpoint networks are perfectly aligned and have learned the same features, then we should have $\mathrm{corr}(X_1, X_2) = 1$. But in practice, it is more typical for pairs of aligned units (whose alignment minimizes the cost function given by \Eqref{eqn:matching}) to have a correlation of $\mathrm{corr}(X_1, X_2) \approx 0.4$. When considering the midpoint interpolated network ($\alpha = 0.5$), the variance of $X_\alpha$ is given by
\begin{align*}
    \Var(X_\alpha) &= \Var\left(\frac{X_1 + X_2}{2}\right) \\
    &= \frac{\Var(X_1) + \Var(X_2) + 2\Cov(X_1, X_2)}{4} \\
    &= \frac{\mathrm{std}^2(X_1) + \mathrm{std}^2(X_2) + 2 \cdot \mathrm{corr}(X_1, X_2) \cdot \mathrm{std}(X_1)\mathrm{std}(X_2)}{4} \\
    &= \left(\frac{\mathrm{std}(X_1) + \mathrm{std}(X_2)}{2}\right)^2 - \frac{(1 - \mathrm{corr}(X_1, X_2))}{2}\mathrm{std}(X_1)\mathrm{std}(X_2).
\end{align*}
We typically have $\mathrm{std}(X_1) \approx \mathrm{std}(X_2)$, so that this simplifies to $\Var(X_\alpha) = (0.5 + 0.5\cdot\mathrm{corr}(X_1, X_2))\cdot\Var(X_1)$. With our typical value of $\mathrm{corr}(X_1, X_2) \approx 0.4$ for aligned networks, this yields $\Var(X_\alpha) = 0.7\cdot\Var(X_1)$: a 30\% reduction compared to the endpoint networks.
This analysis cannot be rigorously extended to deeper layers of the interpolated network, but intuitively we expect this decay to compound with depth. This intuition matches our experiments, where we see that variance collapse becomes worse as we progress through the layers of MLP, VGG, and ResNet50 networks (\Figref{fig:layerwise}).

\section{\name}\label{sec:correction}
We propose two methods for addressing variance collapse. Both aim to correct the statistics of hidden units in the interpolated network. We call these methods \name (REnormalizing Permuted Activations for Interpolation Repair).

Given an interpolated network $\theta_\alpha = (1 - \alpha) \cdot \theta_1 + \alpha \cdot \theta_2$ for some $0 < \alpha < 1$ (with aligned endpoint networks $\theta_1, \theta_2$), we select the set of hidden units or channels whose statistics we aim to correct. For example, for VGG networks we correct the preactivations of every convolutional layer. For ResNets we correct both these convolutional preactivations and the outputs of each residual block.

Our goal will be to compute a set of affine (rescale-and-shift) coefficients for every selected channel, such that the statistics of all selected channels are corrected. Let us consider a particular channel, \eg the 45th convolutional channel of the 8th layer in an interpolated ResNet18. Similar to the analysis in the last section, let $X_1$ and $X_2$ be the values of the channel in the two endpoint networks, viewed as random variables over the input training data, and let $X_\alpha$ be the same channel in the interpolated network. Then we want the following two conditions to hold:
\begin{align}\label{eq:statprop}
    \E[X_\alpha] &= (1 - \alpha)\cdot\E[X_1] + \alpha\cdot\E[X_2], \\
    \label{eq:statprop2}
    \mathrm{std}(X_\alpha) &= (1 - \alpha)\cdot\mathrm{std}(X_1) + \alpha \cdot \mathrm{std}(X_2).
\end{align}
Whereas before any correction, we typically have $\mathrm{std}(X_\alpha) \ll \min(\mathrm{std}(X_1), \mathrm{std}(X_2))$ due to variance collapse. In the following, we present two algorithms to compute the appropriate sets of affine coefficients for each selected channel, in order to induce these conditions.

Both algorithms depend upon the computation of the statistics $\E[X_1],\,\E[X_2],\,\mathrm{std}(X_1),\,\mathrm{std}(X_2)$ for each selected channel in the endpoint networks. We present a PyTorch-based approach to perform this computation  in Appendix~\ref{app:repairpseudo}.

\begin{figure}
    \centering
    \subfigure{\includegraphics[height=3.8cm]{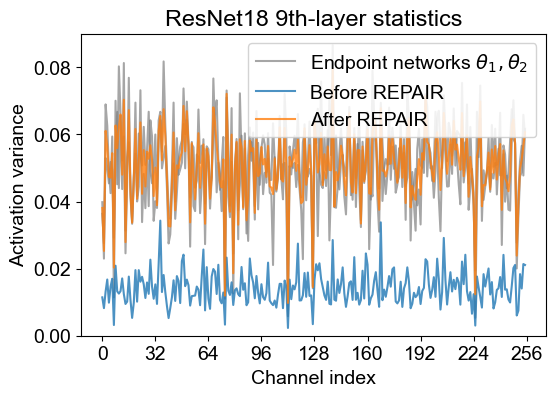}}
    \subfigure{\includegraphics[height=3.8cm]{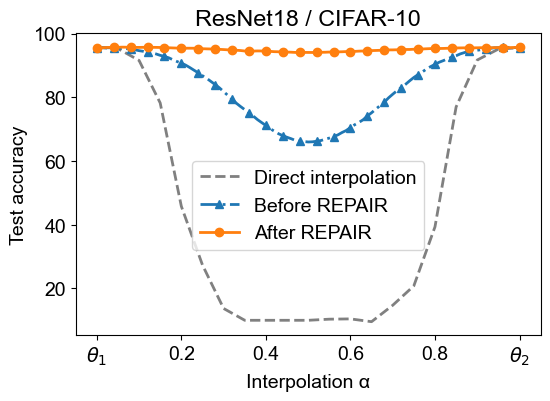}}
    \subfigure{\includegraphics[height=3.8cm]{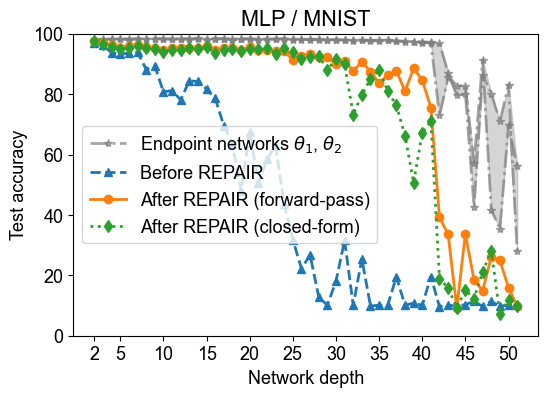}}
    \caption{\small{\textbf{\name restores the internal statistics of averaged neural networks.} \textbf{Left:} We visualize the statistics of different channels in 9th layer of an interpolated ResNet18 on CIFAR-10. The uncorrected network undergoes variance collapse, whereas \name restores the internal statistics of the network to be similar to the endpoint networks. \textbf{Middle:} After applying \name, internal statistics are restored, and barrier is reduced to 1.5\% for CIFAR-10. \textbf{Right:} Permuted interpolation without a statistical correction (before \name) only performs well when limited to MLPs of a few layers \citep{entezari2021role}. \name enables high-performance weight-space averaging between much deeper aligned MLPs.}}
    \label{fig:basicmlpbarrier}
\end{figure}

\subsection{Closed-form approximate variant}
We first present an efficient, approximate algorithm which computes the desired affine coefficients without using forward-passes in the interpolated network. Consider a hidden unit in the first layer of the interpolated network. As before, let $X_\alpha$ represent the unit in the interpolated network, and $X_1, X_2$ the same unit in the two endpoint networks, respectively. Condition~(\ref{eq:statprop}) will already be satisfied for this unit by virtue of the equation $X_\alpha = (1 - \alpha) \cdot X_1 + \alpha \cdot X_2$. Given the values $\Var(X_1), \Var(X_2)$, and $\Cov(X_1, X_2)$, it is possible to compute the variance of $X_\alpha$ exactly according to the formula
\begin{equation}
\label{eq:firstvarform}
    \Var(X_\alpha) = (1-\alpha)^2\Var(X_1) + \alpha^2\Var(X_2) + 2\alpha(1-\alpha)\Cov(X_1, X_2).
\end{equation}
Therefore, to satisfy condition~(\ref{eq:statprop2}) for this unit, the rescaling coefficient $\beta$ must be
\[
\beta = \frac{(1 - \alpha)\cdot\mathrm{std}(X_1) + \alpha \cdot \mathrm{std}(X_2)}{\sqrt{(1-\alpha)^2\Var(X_1) + \alpha^2\Var(X_2) + 2\alpha(1-\alpha)\Cov(X_1, X_2)}},
\]
which is simply the desired standard deviation divided by the standard deviation of $X_\alpha$. For each unit in the first layer, this factor is exactly correct in order to obtain the desired statistics. For deeper layers, this factor is an approximation which we empirically test.

In \Figref{fig:basicmlpbarrier} (right), we apply this rescaling to every hidden unit of MLPs of depth between 2 and 50 hidden layers, which are linearly interpolated ($\alpha = 0.5$) between aligned endpoints networks trained on MNIST. We find that this rescaling significantly improves the performance of such interpolated networks. In particular, we obtain interpolated checkpoints of up to 27 layers that achieve over 90\% accuracy, whereas without a correction, we hit this limit after only 6 layers.

We note that this correction requires forward passes in the endpoint networks in order to compute the values of $\Var(X_1), \Var(X_2)$, and $\Cov(X_1, X_2)$ for each hidden unit. Once these values have been computed, correction coefficients to interpolated networks across arbitrary choice of interpolation coefficient $\alpha$ can be generated without requiring any further forward passes.

\subsection{Forward-pass exact variant}
The rescaling coefficients generated by the above algorithm are approximate, only being guaranteed to induce the desired conditions~(\ref{eq:statprop}), (\ref{eq:statprop2}) for hidden units or channels in the first layer of an interpolated network. We find that it is effective for the case of deep MLPs, but in our experiments it was insufficient to significantly reduce the interpolation barrier for more challenging cases like ResNet50 trained on ImageNet. We now propose an exact algorithm, which uses forward passes in the interpolated networks in order to generate affine (rescale-and-shift) parameters for every channel in the network that we aim to correct. This method outperforms the approximate variant, especially for challenging cases.

The exact method proceeds as follows. For each module in the interpolated network whose outputs we have identified as targets for statistical correction, we apply a wrapper, PyTorch pseudocode for which can be found in Appendix~\ref{app:repairpseudo}. This wrapper adds a Batch Normalization layer after the wrapped module which is initially set to ``train'' mode. Each such added BatchNorm layer contains affine (\ie per-channel rescale-and-shift) parameters. For a given channel incoming to the BatchNorm layer, we set the respective affine weight to $(1 - \alpha)\,\mathrm{std}(X_1) + \alpha\,\mathrm{std}(X_2)$ and the bias to $(1-\alpha)\,\E[X_1] + \alpha\,\E[X_2]$, where $X_1$ and $X_2$ are the respective channels in the endpoint networks as in conditions~(\ref{eq:statprop}), (\ref{eq:statprop2}).

With the added BatchNorm layers in training mode, these affine parameters will exactly induce our statistical conditions with respect to batches of input data. The reason for this is that during execution, the added BatchNorm layers first renormalize their inputs to have zero mean and unit variance per channel, and then apply our given affine transformation which sets the statistics of the output to be that of conditions~(\ref{eq:statprop}), (\ref{eq:statprop2}). Next, we pass a set of training data through the network ($\sim$5,000 examples suffices) so that the running mean and variance parameters of our added BatchNorm layers will be accurately estimated. During this pass, any BatchNorm layers which already existed in the original network are kept frozen. Finally, we set the added BatchNorm layers to evaluation mode, so that they behave as affine layers which do not recompute statistics. At this point, the resulting network is functionally equivalent to one in which the weights of our selected set of channels have been rescaled and biases shifted. If we wish to generate a new parameter vector $\theta_\alpha'$ which is compatible with the original network architecture (\ie lacks these added BatchNorm layers), then we can perform BatchNorm layer fusion~\citep{markus2018fusing} in which appropriate rescaling and bias-shifting values are computed from each added BatchNorm layer, and then applied to the preceding convolutional filters.

The networks resulting from this process have their internal statistics corrected, so that all selected channels satisfy conditions~(\ref{eq:statprop}), (\ref{eq:statprop2}). In \Figref{fig:basicmlpbarrier} (left) we observe that \name has resolved variance collapse. In \Figref{fig:basicmlpbarrier} (middle) we apply \name to networks where the weights are linearly interpolated between a pair of  ResNet18s whose hidden units have been brought into alignment. Before the correction, networks near the midpoint have a reduced accuracy of 66.0\% on the CIFAR-10 test set, while the endpoints accuracy is 95.5\%. After correction, all checkpoints along the linear path have significantly boosted accuracy, with the midpoint performing at 94.1\%. In comparison, \citet{singh2020model} report a linear midpoint accuracy of 77.0\% using strong optimal-transport based alignment methods to improve over the baseline.

Throughout the rest of the paper, we refer to the above method as \name. We note that the utility of this method is orthogonal to any improvements in the algorithm used to align the hidden units of the endpoint networks. We explore this in Appendix \Figref{fig:without_alignment}, where we report barrier curves for \name applied to interpolations between unaligned networks as well. In the remainder of the paper, we demonstrate the effectiveness of \name across a wide range of scenarios.

\section{Experimental results}
\label{sec:results}
\begin{figure}
    \centering
    \subfigure{\includegraphics[height=3.7cm]{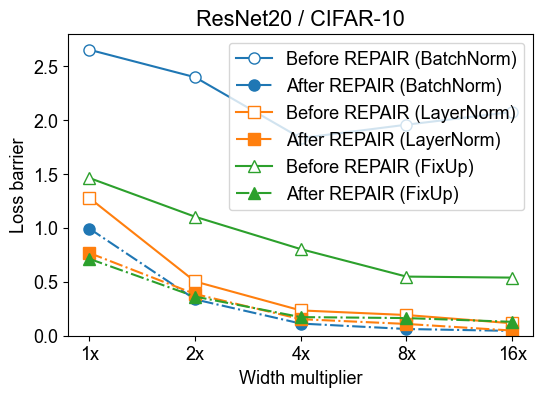}}
    \hspace{1em}
    \subfigure{\includegraphics[height=3.7cm]{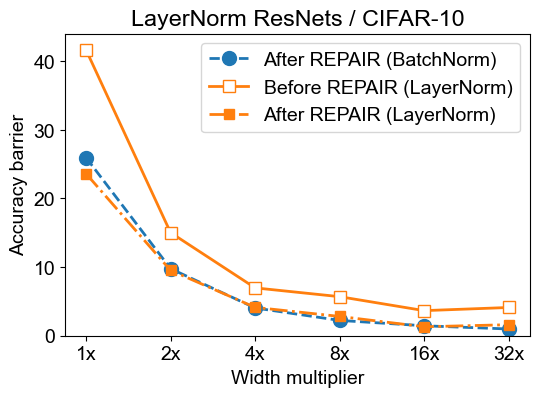}}
    \subfigure{\includegraphics[height=3.7cm]{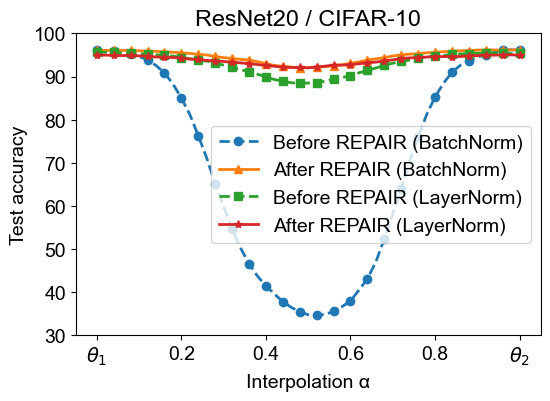}}
    \caption{\small\textbf{Effect of normalization layer.} \textbf{Left:} Loss barriers with and without \name for ResNet20s trained with BatchNorm, LayerNorm, and normalization-free via FixUp, varying the width multiplier from 1 to 16. \textbf{Middle:} LayerNorm networks are unique in reaching a relatively low barrier before \name. \textbf{Right:} Performance curves of networks interpolated between aligned ResNet20s. Without \name, the midpoint BatchNorm-based network achieves 34.7\% accuracy, compared to 88.4\% for LayerNorm. After \name, both variants attain 92.0\% accuracy.
    }
\label{fig:normlayer}
\end{figure}

\begin{figure}
    \centering
    \subfigure{\includegraphics[height=5cm]{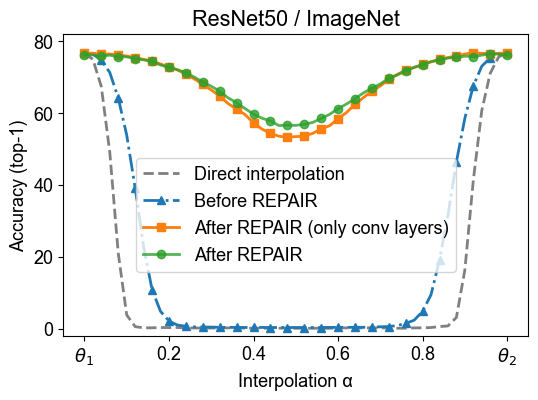}}
    \hspace{0.5cm}
    \subfigure{\includegraphics[height=5cm]{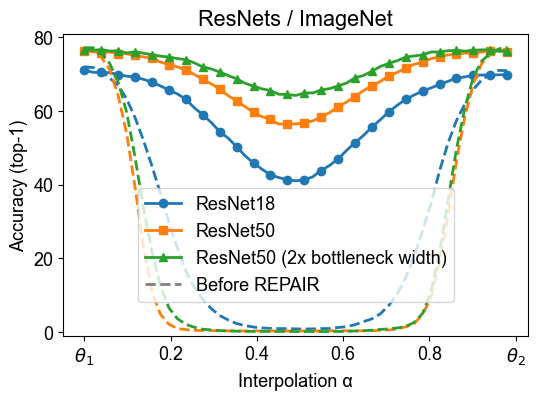}}
    \caption{\small \textbf{\name significantly reduces barrier between ResNet50s trained on ImageNet.} \textbf{Left:} Without \name, interpolations between two aligned, independently trained ResNet50s attain less than 1\% test accuracy on ImageNet. After applying \name to convolutional layer outputs, the midpoint is boosted to 53.2\%. Using full \name, which is also applied to the outputs of residual blocks, this is further boosted to 56.5\%. \textbf{Right:} Larger and wider ResNet architectures have smaller barrier. Dashed lines indicate the baseline of interpolation between aligned networks~\citep{singh2020model,ainsworth2022git}, and solid lines refer to  \name on top of baseline.}
    \label{fig:imageresnet}
\end{figure}

In this section we report the results of the following experiments.
\begin{itemize}
    \item We investigate the effectiveness of \name for three variants of ResNet trained on CIFAR-10: standard BatchNorm-based ResNets, nonstandard LayerNorm-based ResNets, and normalization-free ResNets using FixUp~\citep{zhang2019fixup}. (Section~\ref{sec:normalize})
    \item We apply \name to (interpolations between aligned) ResNets trained on ImageNet. (Section~\ref{sec:imagenet})
    \item We study the relationship between width and barrier size, with experiments conducted using 10-layer MLPs and standard ResNet20s across four datasets: MNIST, SVHN, FashionMNIST, and CIFAR-10. We also report barriers for VGG networks of varying depth trained on CIFAR-10. (Section~\ref{sec:wd})
    \item Finally, we perform a replication using \name of an experiment from \citet{ainsworth2022git}, in which a pair of models trained on disjoint subsets of data is constructively merged. (Section~\ref{sec:sdt})
\end{itemize}
Extra figures which replicate these experiments in terms of alternate metrics can be found in Appendix~\ref{app:moreplots}.

\subsection{Normalization layer} \label{sec:normalize}
We begin by considering ResNet20s trained on CIFAR-10 using three different types of normalization layer: BatchNorm~\citep{ioffe2015batch}, LayerNorm~\citep{ba2016layer}, and normalization-free via Fixup~\citep{zhang2019fixup}. We combine this ablation with a study on the effect of width: for each class of networks, we vary the width multiplier from $1\times$, where the final block has 64 channels, up to $16\times$ or 1024 channels. In \Figref{fig:normlayer} we report the barriers to interpolation, both with and without \name. We find that \name is effective for all three cases, shrinking the loss barrier to below 0.05 for BatchNorm and LayerNorm-based networks when width is scaled to $16\times$.

The LayerNorm-based ResNets are a special case that deserves comment. We first note that LayerNorm is nonstandard in ResNets, and has worse performance when compared to BatchNorm; with our training configuration, LayerNorm-based ResNet20s achieve 92.4\% CIFAR-10 test set accuracy, vs. 93.6\% for the BatchNorm baseline. But these networks also have a unique property: when interpolating between aligned LayerNorm-ResNets, the midpoints already perform relatively well before \name is applied. In \Figref{fig:normlayer}, we observe that the midpoint between our pair of $8\times$-width LayerNorm-ResNet20s attains 90.3\% test-set accuracy, which is boosted to 93.2\% by an application of \name. In contrast, for the same width using BatchNorm, the interpolated network obtains only 32.3\% accuracy, which is boosted to 94.4\% by \name.

One possible explanation for this phenomenon is that LayerNorm can partially mitigate variance collapse because it applies test-time normalization, which may prevent per-layer reductions in variance from compounding through the forward pass. In contrast, BatchNorm is functionally just an affine layer during test-time, so that variance reductions do compound.

We claim that this observation regarding LayerNorm-based ResNets explains the contradiction between the results of \citet{singh2020model} and \citet{ainsworth2022git}. In the former, the authors develop a strong optimal-transport based method of aligning networks, which contains the approach of \citet{li2015convergent} as a special case. They show that even this method is insufficient to attain low-barrier connectivity between standard ResNets before fine-tuning; their best result is a barrier of 16\% error between aligned ResNet18s trained on CIFAR-10. In contrast, \citet{ainsworth2022git} report that a variety of alignment methods, including the method of \citet{li2015convergent} which they call ``activation matching'', suffice to establish nearly zero-barrier connectivity between wide ResNets trained on CIFAR-10. How can both results be true? We claim that this contradiction is resolved by the observation that \citet{ainsworth2022git} replace standard BatchNorm layers with LayerNorm in their ResNets, as is evident in the code release\footnote{https://github.com/samuela/git-re-basin/blob/main/src/resnet20.py\#L18}. In our experiments, the use of LayerNorm uniquely allows for low-barrier linear connectivity without the requirement of a statistical correction such as \name, at the cost of reduced performance on most datasets.

\subsection{ImageNet} \label{sec:imagenet}
Next, we explore the impact of \name on the barriers to interpolation between standard ResNet models trained from scratch on ImageNet~\citep{deng2009ImageNet}. We test ResNet18, ResNet50, and a double-width variant of ResNet50 in \Figref{fig:imageresnet} (right). Without \name, the interpolated midpoints between each aligned pair of networks perform at below 1\% accuracy on the ImageNet validation set. After \name, the midpoint ResNet18 improves to 41.1\%, ResNet50 to 56.5\%, and double-width ResNet50 to 64.2\%.

We find in \Figref{fig:imageresnet} (left) that it is important to apply \name not just to all convolutional layer outputs, but also to the outputs of every residual block. For the case of ResNet50, using \name with these extra channels boosted the performance of the midpoint from 53.2\% to 56.5\%, which is a 14\% reduction in the size of the barrier (from 23.4\% to 20.1\%). We note that for architectures which contain BatchNorm after every convolutional layer, applying \name only to convolutional layer outputs is mathematically equivalent to resetting the BatchNorm statistics of the network. Performing such a reset on averaged networks goes back to \citep{izmailov2018averaging}, but as far as we are aware, has not been applied to interpolation between independently trained networks until now. We provide a comparison between the ImageNet results of our work and those of \citet{ainsworth2022git} in Appendix \Figref{fig:vsrebasin}.

In general, the barriers for these architectures on ImageNet are still relatively high. The standard ResNet50 architecture has a final-block bottleneck width of 1024, and we measure the barrier after \name to be 20.1\% in terms of test error. The double-width ResNet50 variant has a final-block bottleneck width of 2048, reducing the  barrier to 12.9\%. In comparison, the widest ResNet20 we studied on CIFAR-10 had final-layer width of 1024, and a barrier of nearly zero. Therefore, it may be the case that for more difficult datasets, larger widths are required in order to reach low barriers.

\begin{figure}
    \centering
    \subfigure{\includegraphics[height=3.9cm]{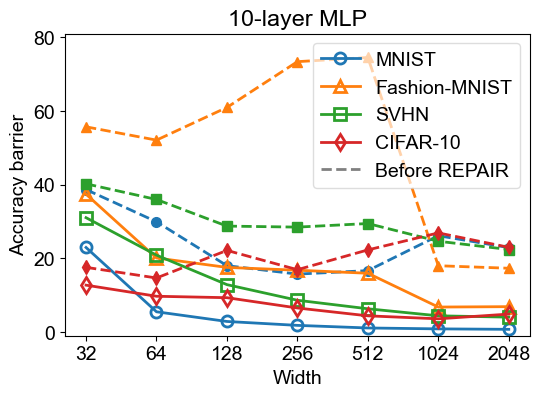}}
    \subfigure{\includegraphics[height=3.9cm]{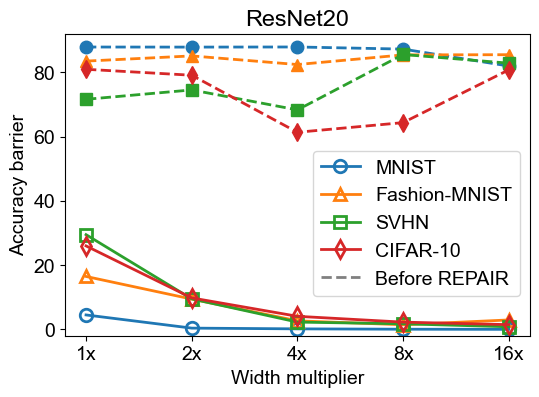}}
    \subfigure{\includegraphics[height=3.9cm]{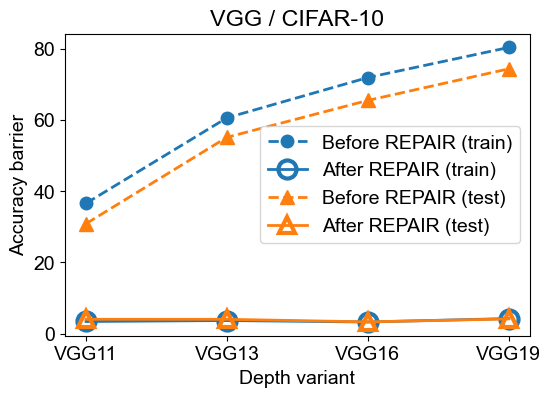}}
    \caption{\small{\textbf{Network width and depth.} \textbf{Left:} We investigate the effect of \name on the barrier for 10-layer MLPs, trained on MNIST, FashionMNIST, SVHN, and CIFAR-10. In each case, the baseline (interpolation between aligned networks) is shown with  dotted curve, and the solid curve refers to cases where \name is applied on top of the baseline. We vary the width from 32 to 2048 hidden units per layer. \textbf{Middle:} We conduct the same experiment using ResNet20s trained on the same four datasets. We vary the width multiplier from 1 to 16, producing models whose final block ranges from 64 to 1024 channels. \textbf{Right:} We investigate the effect of \name on the  barrier for VGG networks trained on CIFAR-10. We vary the network depth from 11 to 19 layers, and observe that without \name the barrier increases with depth, whereas with \name it is close to constant in depth.}}
    \label{fig:fourdatasets}
\end{figure}

\subsection{Network width and depth} \label{sec:wd}
In \Figref{fig:fourdatasets} (left and middle), we study the impact of our statistical correction across four datasets (MNIST, FashionMNIST~\citep{xiao2017fashion}, SVHN~\citep{netzer2011reading}, and CIFAR-10) and two architectures (10-layer MLP and standard ResNet20). We report the barriers to interpolation in terms of accuracy both before and after \name, varying the width of each network. In all cases, \name is effective in reducing the barrier, and the size of the barrier decreases as network width is increased. For MNIST, which is the easiest task, even networks of smaller width are able to reach nearly zero barrier to interpolation.

We also study the effect of network depth, across standard VGG networks of depth varying from 11 to 19 layers. We report the barriers to interpolation between such aligned networks, before and after \name, in \Figref{fig:fourdatasets} (right). Consistently with our earlier results on MLPs, we find that the barrier increases with depth before \name is applied, and afterwards the barrier is small and roughly flat with depth.

\subsection{Split data training} \label{sec:sdt}
\begin{wrapfigure}{r}{0.45\textwidth}
\centering
\includegraphics[width=0.42\textwidth]{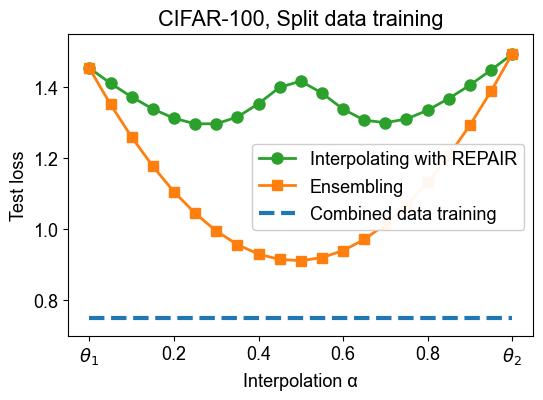}
    \caption{\small \textbf{Split data training.} When two networks are trained on disjoint, biased subsets of CIFAR-100, their {\name}ed interpolations outperform either endpoint with respect to the combined test set.}
\label{fig:cifar100}
\end{wrapfigure}

In this section we study the setting where two endpoint networks are trained on disjoint splits of the training dataset. We aim to replicate the corresponding experiment of \citet{ainsworth2022git}, but using \name applied to standard BatchNorm networks. 

The experiment proceeds as follows. We first split the CIFAR-100 training set, consisting of 50,000 images distributed across 100 classes, into two disjoint sets of 25,000 images. The first split to consists of a random 80\% of the images in the first 50 classes, and 20\% of the images in the second 50 classes, with the second split having the proportions reversed. We then train two networks, one for each split. The result is that one network is more accurate on the first 50 classes, and the other more accurate on the second, with both performing worse than either their ensemble or a network trained on the full training set.

We next align the hidden units of these two networks, and generate a series of linearly interpolated networks between the two, applying \name to each (\Figref{fig:cifar100}). We find that many of these interpolated networks significantly outperform either of the two endpoints in terms of loss on the full CIFAR-100 test set. In this sense, the endpoint networks can be said to have been constructively merged. The best interpolated network of \citet{ainsworth2022git} was reported to obtain a loss of 1.73 at $\alpha \approx 0.3$. Using \name, our best interpolated network achieves a test loss of 1.30, also with mixing coefficient $\alpha = 0.3$. We attribute this improvement partially to \name, and partially to the increased performance of standard ResNets compared to the LayerNorm-based variants used in \citet{ainsworth2022git}. In Appendix~\ref{app:moreplots}, we compare these results against a strong baseline.

\section{Discussion and future work}
In this paper we proposed \name, a method of mitigating the \emph{variance collapse} phenomenon which we show occurs in interpolated networks. We demonstrated that \name significantly improves the performance of interpolated networks across a wide variety of architectures and datasets. For example, we used \name to reduce the barrier to permuted interpolation for a standard ResNet18 trained on CIFAR-10 from 16\%~\citep{singh2020model} to 1.5\%. Further, we use \name to improve the performance of interpolated ResNet50s from below 1\% to 56.5\% on ImageNet. \name is effective for networks trained without normalization layers, or with LayerNorm instead of BatchNorm.

To explain these results, we provided an analysis of the variance collapse phenomenon and how \name mitigates it. We also demonstrated that LayerNorm-based networks are unique in attaining a relatively low barrier to aligned interpolation before the use of \name, resolving the contradiction between the results of \citet{singh2020model} and \citet{ainsworth2022git}.

In so far as we establish low-barrier permuted interpolation for further scenarios, these results provide support for the conjecture of \citet{entezari2021role}. On the other hand, to practically do so we needed to rescale the preactivations of the interpolated network, moving us out of the realm of strictly permuted linear interpolation. Our rescaling \name evolved as a generalization of the method of resetting BatchNorm statistics of averaged networks~\citep{izmailov2018averaging}. Such a correction appears to be necessary in order to establish low-barrier permuted linear connectivity, at practical widths and without the use of LayerNorm.

On the empirical side, we demonstrated that {\name}ed interpolations between two networks which were trained on disjoint, biased dataset splits are able to outperform either network from which they were interpolated. We hope that further such applied results will be possible, including improvements to weight-space ensembling, checkpoint averaging and robust finetuning.

\section*{Acknowledgements}
We thank Luke Johnston for his insightful comments on the manuscript.

\bibliographystyle{apalike}
\bibliography{references.bib}
\newpage
\appendix
\section*{Appendix}
\section{Further discussion of related work}\label{app:further_related}
\citet{entezari2021role} try to find a low-barrier permutation between two SGD solutions by searching in the set of all permutations using simulated annealing. Neuron alignment methods use efficient heuristics to find a matching that maximizes a defined similarity measure given two neural networks ~\citep{li2015convergent,ashmore2015method,singh2020model, tatro2020optimizing, pittorino2022deep}. This similarity measure is often based on correlation between weights or activations of the neurons in the same layer. Prior work was unable to achieve low-barrier for many standard architectures and/or challenging tasks (\eg ImageNet classification).


\citet{uriot2020safe} use the neuron representation introduced in~\citet{li2015convergent} to perform alignment between different feedforward networks (which they refer to as a crossover between parents).
They consider two ways of characterizing the relationships between neurons: Canonical Correlation Analysis~\citep{hotelling1992relations}\footnote{Canonical Correlation Analysis is a multivariate statistical technique that finds maximally correlated linear relationships
between two sets of observations, under orthogonality and norm constraints.} and
pairwise cross-correlation~\citep{li2015convergent}, and apply alignment operators depending on the uncovered relationship. \citet{he2018multi} propose a neuron alignment method in the context of multi-task model compression. The algorithm leverages layer-wise Hessian approximation to match neurons by computing a similarity measure based on their functional difference. \citet{singh2020model} propose a model fusion algorithm which leverages optimal transport to perform neuron alignment in the Wasserstein space~\citep{agueh2011barycenters}, achieving a barrier of 16\% for ResNet18 trained on CIFAR-10; furthermore they show that by finetuning such interpolated networks, the original performance is recovered.

\citet{tatro2020optimizing} point out that special care must be taken for networks that contain batch normalization layers when it comes to neuron alignment. They match neurons by optimizing a curve in the weight space between two models by aligning correlations of post-activations. The running statistics are normalized by training the model for one epoch, while freezing all learnable parameters of the model. However, the crucial role of statistics reset in the context of linear interpolation of the models is not investigated. A recent work by \citet{pittorino2022deep} uses LayerNorm as part of symmetry removal, so that the network behavior remains unchanged. 

\section{Implementation details} \label{app:implementationdetails}
\subsection{Training hyperparameters}
Table \ref{train_params} summarizes the hyperparameters we used to train the neural networks which appear in this work.
\begin{itemize}
    \item We train all networks using SGD with momentum 0.9. The weight decay and learning rates differ for each task, and are specified below.
    \item For our MLP trainings, we keep the hyperparameters below constant across varying widths, depths, and datasets.
    \item When training on MNIST and SVHN, we remove cutout and horizontal flip from the list of data augmenations used by our ResNet20 training. Otherwise, we keep the below ResNet20 hyperparameters constant across different choices of width, dataset, and normalization layer.
\end{itemize}

\begin{table}[h]
  \caption{Training hyperparameters}
  \label{train_params}
  \centering
  \begin{threeparttable}
  \begin{tabular}{lllll}
    \toprule
    Hyper-parameters     & MLP          & VGG        & ResNet20           & ResNet50/ImageNet \\
    \midrule
    Batch Size        & 2000  & 500    & 500           & 512     \\
    Epochs            & 100   & 100    & 200  & 300 \\
    Learning Rate     & Linear 0.2     & Cosine 0.08   & Cosine 0.4   & Linear 0.5     \\
    Weight decay      & 0.0   & 0.0005 & 0.0001 & 0.0001  \\
    Data augmentation & Translate & Flip+Translate & Flip+Translate+Cutout & Flip+RandomResizedCrop \\
    
    \bottomrule
  \end{tabular}
  \end{threeparttable}
\end{table}

\subsection{PyTorch wrapper module pseudocode}\label{app:repairpseudo}
\begin{enumerate}
    \item The first step of \name is to measure the statistics of identified channels in the endpoint networks. There are many ways to do this, but we found that the following was efficient and had low code-complexity in a Pytorch environment. For each module in the interpolated network whose outputs we wish to \name, we wrap the corresponding modules in the endpoint networks with the following:
    \begin{lstlisting}[language=python]
class TrackLayer(nn.Module):
    def __init__(self, layer):
        super().__init__()
        self.layer = layer
        self.bn = nn.BatchNorm2d(len(layer.weight))
        self.bn.train()
        self.layer.eval()
    def get_stats(self):
        return (self.bn.running_mean, self.bn.running_var.sqrt())
    def forward(self, inputs):
        outputs = self.layer(inputs)
        # Apply BatchNorm so that the running mean/variance are updated; discard the output.
        self.bn(outputs)
        return outputs
    \end{lstlisting}
    It then suffices to pass a small set of training data ($\sim$5,000 examples) through the tracked endpoint networks. At this point, the statistics of each tracked module's output can be retrieved from the wrapping TrackLayer.
    \item Next, we wrap each module in the interpolated network that we wish to \name with the following:
    \begin{lstlisting}[language=python]
class ResetLayer(nn.Module):
    def __init__(self, layer):
        super().__init__()
        self.layer = layer
        self.bn = nn.BatchNorm2d(len(layer.weight))
    def set_stats(self, goal_mean, goal_std):
        self.bn.bias.data = goal_mean
        self.bn.weight.data = goal_std
    def forward(self, x):
        return self.bn(self.layer(x))
    \end{lstlisting}
    The next step is to iterate over each triple of corresponding (TrackLayer, TrackLayer, ResetLayer) modules coming from the two endpoint networks and the interpolated network. For each triple, we set the statistics of the ResetLayer to be the interpolation of the statistics from the two TrackLayers. Working and minimal code to accomplish this can be found in our code release.
\end{enumerate}

\section{Additional plots} \label{app:moreplots}

\begin{figure}[h]
    \centering
    \subfigure{\includegraphics[height=5cm]{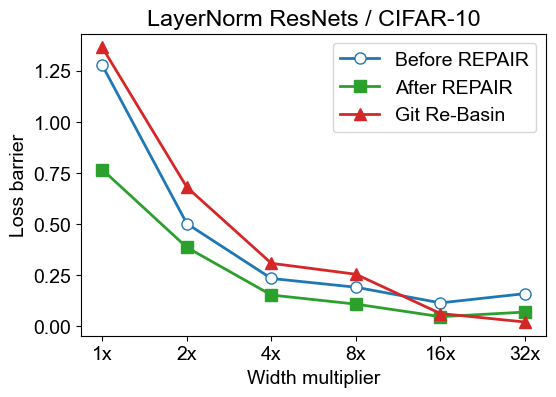}}
    \hspace{0.5cm}
    \subfigure{\includegraphics[height=5cm]{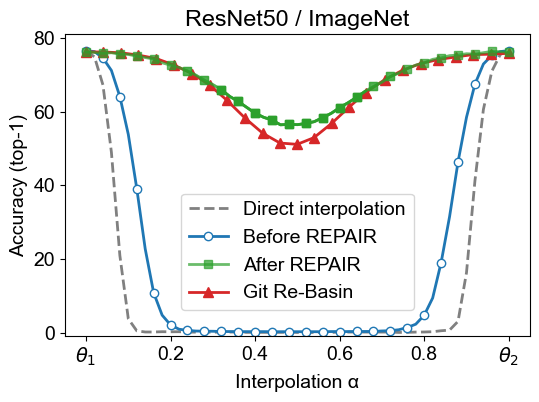}}
    \caption{\small{\textbf{Comparison to \citet{ainsworth2022git}.} \textbf{Left:} We compare barrier values of LayerNorm-based ResNets, for our baseline and \name, to the barriers reported in \citet{ainsworth2022git}. We find that for networks of width up to 8$\times$, our baseline barrier values are comparable, indicating that our baseline alignment method \citep{li2015convergent} performs similarly to those explored in \citet{ainsworth2022git}. With \name, our barrier values are lower. For the case of 32$\times$-width networks, our barrier value is higher than that reported in \citet{ainsworth2022git}. \textbf{Right:} We compare our results on ImageNet using \name with those of \citet{ainsworth2022git}. We note that in their ImageNet experiments, \citet{ainsworth2022git} do reset the BatchNorm statistics in their interpolated ResNet50s based on our suggestion to do so$^3$. The extra barrier reduction in our results can therefore be attributed to the fact that \name also statistically corrects the outputs of each residual block.}}
\label{fig:vsrebasin}
\end{figure}
\footnotetext[3]{See \url{https://twitter.com/kellerjordan0/status/1570837651741364226} and Section A.3.3 of \citet{ainsworth2022git} \url{https://arxiv.org/abs/2209.04836v3}.}

\begin{figure}
    \centering
    \subfigure{\includegraphics[width=0.32\textwidth]{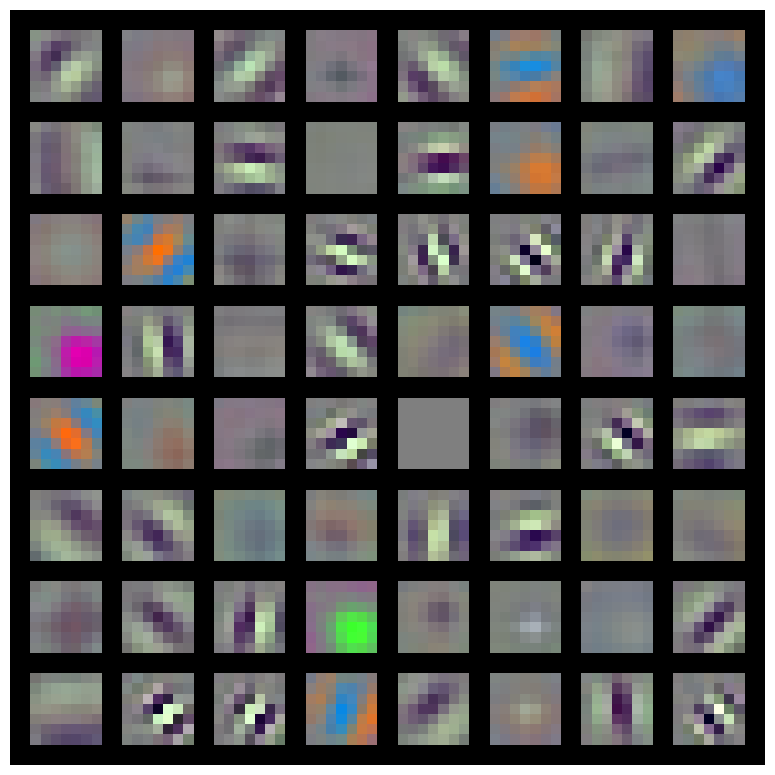}}
    \subfigure{\includegraphics[width=0.32\textwidth]{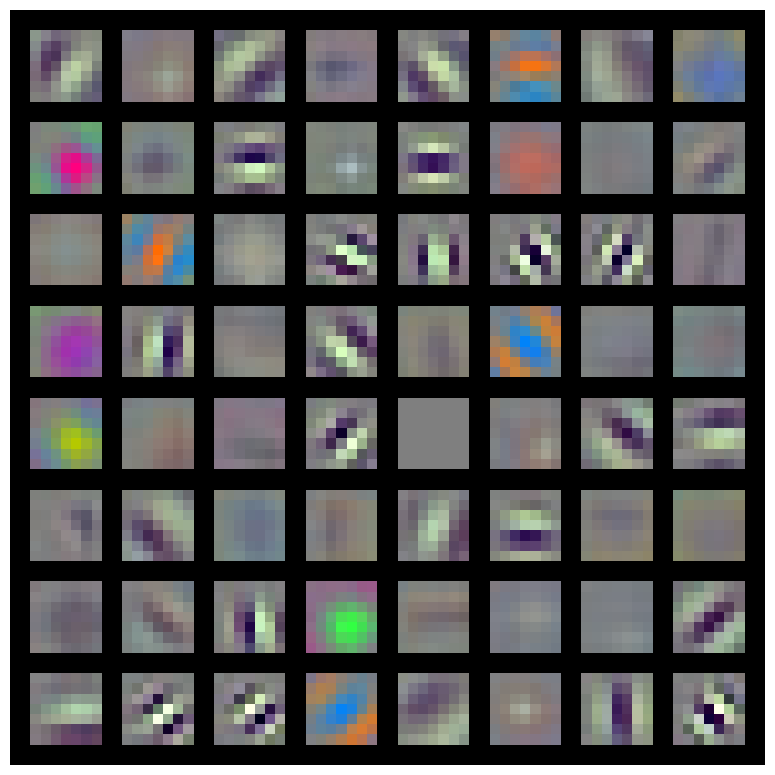}}
    \subfigure{\includegraphics[width=0.32\textwidth]{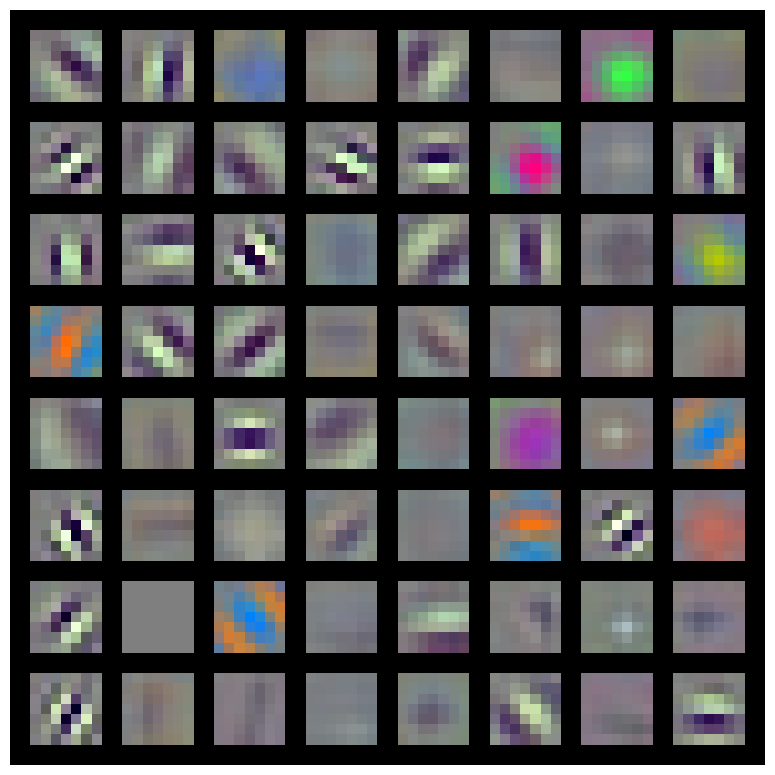}}
    \caption{\small \textbf{Aligned convolutional filters for the first layer in ResNet50.} We train two ResNet50s on ImageNet independently, calling these models A and B. \textbf{Left:} The first-layer convolutional filters of model A. \textbf{Middle:} The filters of model B, having been permuted in order to maximize the total activation-wise correlation to the corresponding filters of model A. Some paired filters appear nearly identical, while for others there was no close match. \textbf{Right:} The filters of model B in their original positions.}
    \label{fig:resnet50_filts}
\end{figure}

\begin{figure}
    \centering
    \subfigure{\includegraphics[height=5cm]{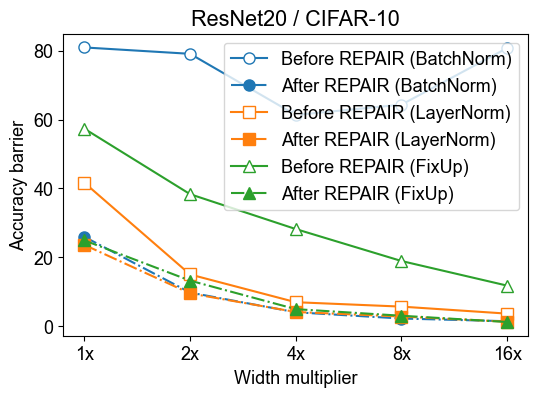}}
    \hspace{0.5cm}
    \subfigure{\includegraphics[height=5cm]{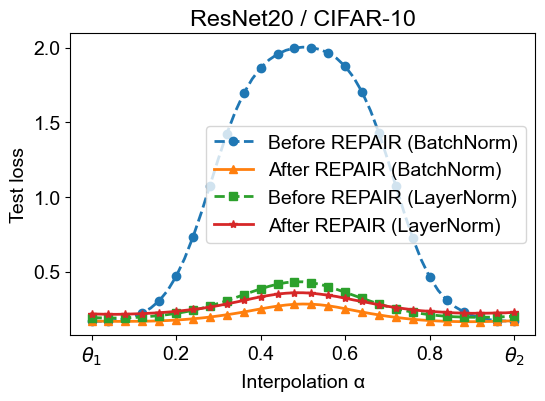}}
    \caption{\small{\textbf{Effect of normalization layer.} \textbf{Left:} We replicate \Figref{fig:normlayer} (left) with barriers measured in terms of test accuracy. \textbf{Right:} We replicate  \Figref{fig:normlayer} (right) with barriers measured  in terms of test loss.}}
\label{fig:normlayer2}
\end{figure}

\begin{figure}
    \centering
    \subfigure{\includegraphics[height=3.5cm]{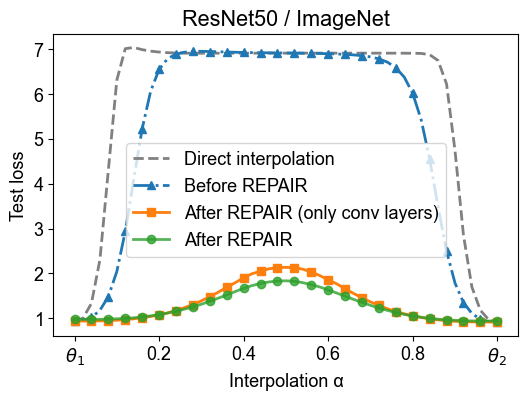}}
    \subfigure{\includegraphics[height=3.5cm]{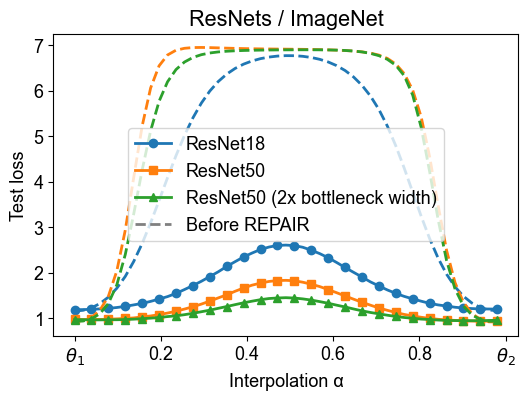}}
    \subfigure{\includegraphics[height=3.5cm]{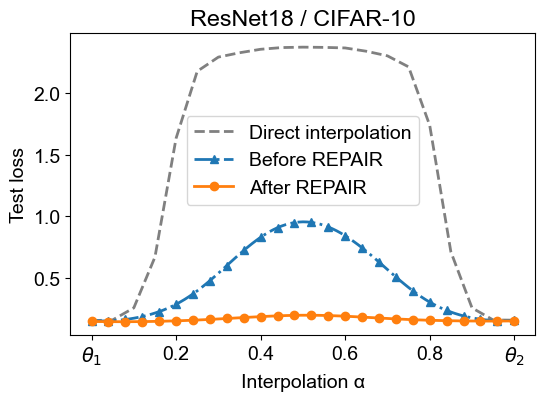}}
    \caption{\small{\textbf{Barrier curves in terms of loss.}  We report the performance of interpolated ResNets in terms of test loss on ImageNet and CIFAR-10. The corresponding figures in terms of accuracy are \Figref{fig:imageresnet} and \Figref{fig:basicmlpbarrier} (middle)}.}
\label{fig:imagenet2}
\end{figure}

\begin{figure}
    \centering
    \subfigure{\includegraphics[height=3.8cm]{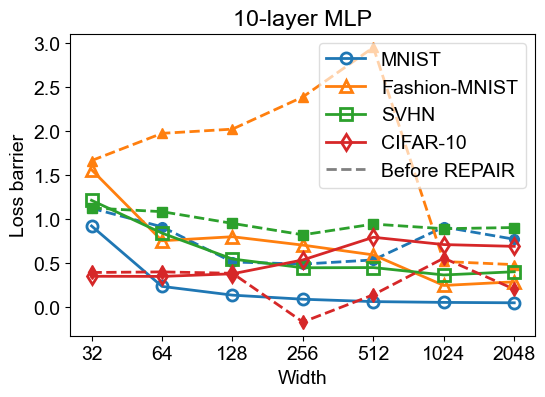}}
    \subfigure{\includegraphics[height=3.8cm]{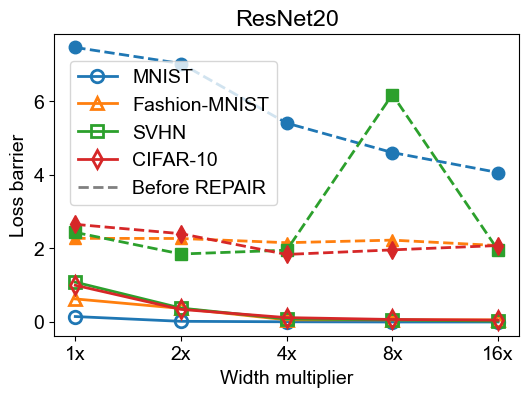}}
    \subfigure{\includegraphics[height=3.8cm]{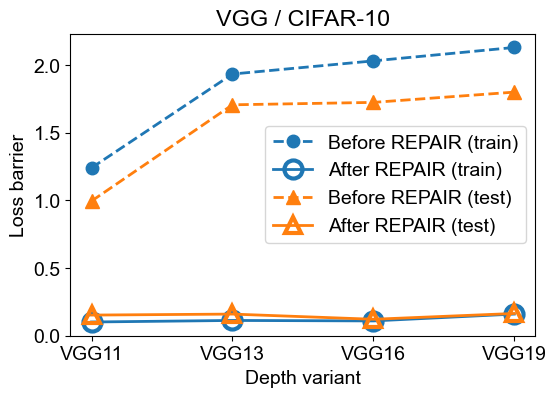}}
    \caption{\small{\textbf{Network width and depth.} We replicate \Figref{fig:fourdatasets} with barriers measured in terms of test loss instead of test accuracy.}}
    \label{fig:fourdatasets2}
\end{figure}

\begin{figure}
    \centering
    \subfigure{\includegraphics[height=5cm]{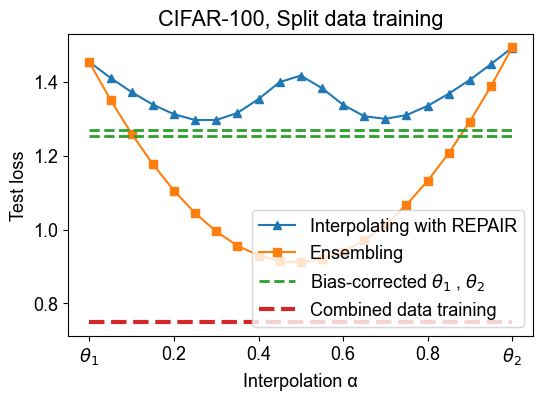}}
    \hspace{0.6cm}
    \subfigure{\includegraphics[height=5cm]{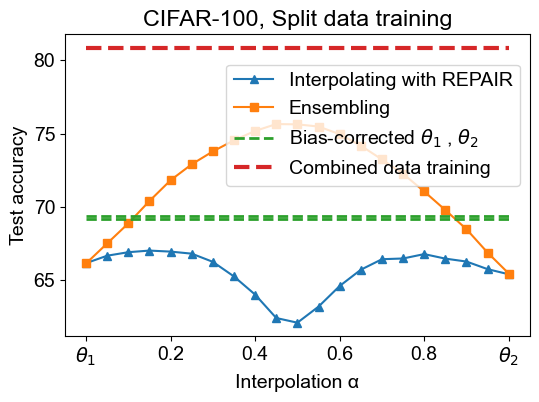}}
    \caption{\small\textbf{CIFAR-100 split data experiment against a baseline.} Two networks are independently trained on disjoint, biased subsets of the CIFAR-100 training set. With respect to performance on the full CIFAR-100 test set, we report the performance of ensembling/mixing the logit outputs of the two models (orange), and the performance of interpolations in weight-space between two aligned models (blue), with \name applied to every interpolated network. As baselines, we report the performance of a single model which was trained on the full CIFAR-100 training set (red), and the performance of the two endpoint networks, after a distribution-shift correction has been applied (green). We find that there exist {\name}ed interpolated checkpoints which outperform either endpoint network in terms of both test loss and accuracy, showing that constructive merging of models is possible. The distribution-shift correction (green line) is as follows. Consider an endpoint network A; it was trained with a training set that was biased 4:1 towards the first 50 classes, so that on the test set, network A over-predicts these first 50 and under-predicts the second 50 classes, leading to increased loss. Our baseline is then to scale down the first 50 logits of network A such that its predictions on the test set become balanced over the 100 classes. We find that this baseline, applied separately to either endpoint network (thus two green lines), outperforms the best gain that can be obtained from constructive merging of the two models via {\name}ed interpolation.}
\label{fig:cifar100app}
\end{figure}

\begin{figure}
    \centering
    \subfigure{\includegraphics[width=5cm]{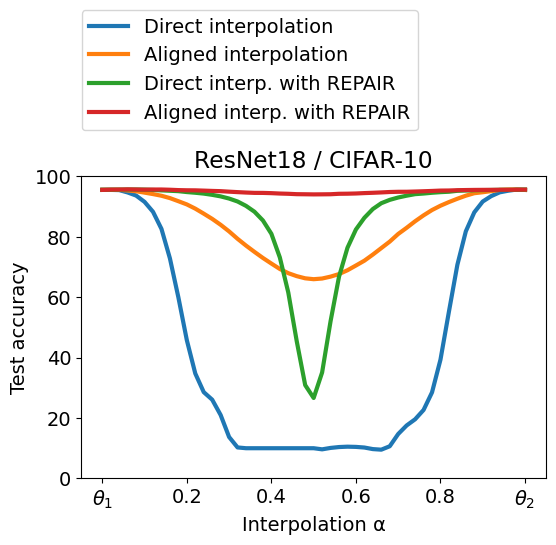}}
    \subfigure{\includegraphics[width=5cm]{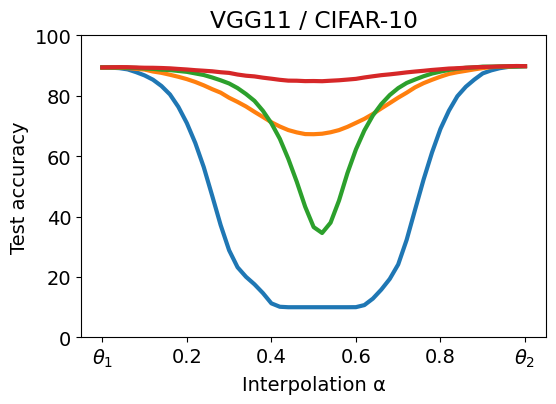}}
    \subfigure{\includegraphics[width=5cm]{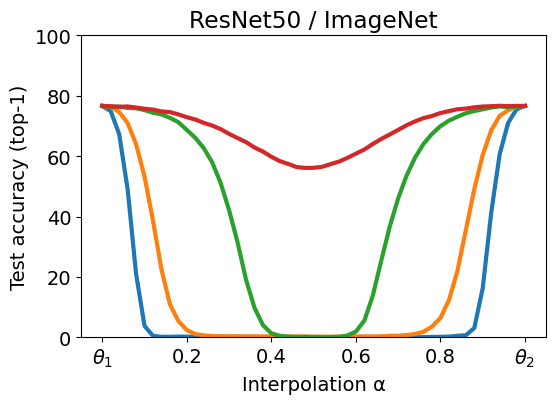}}
    \caption{\small{\textbf{\name without alignment.} In the preceding experiments, we have always first aligned the neurons of the two endpoint networks before interpolating. In this figure we also report accuracy curves for interpolations between the original unaligned endpoint networks, both with and without \name. We note that in the case of ResNet18, we apply \name to convolutional outputs, which is exactly equivalent to resetting BatchNorm statistics. For VGG11, the architecture does not contain normalization layers, and \name is applied to convolutional outputs. For ResNet50, we apply \name to both convolutional outputs and residual block outputs, as described in \Secref{sec:imagenet}. It appears that \name is effective towards the endpoints, and neuron-alignment becomes essential near the midpoint. Both methods are necessary in order to have a high-performing midpoint.}}
    \label{fig:without_alignment}
\end{figure}

\end{document}